# Potential of Artificial Intelligence Algorithms for Identification of Relevant Diagnostic and Prognostic Biomarkers of Early-Stage Liver Cancer

Ali Bou Nassif*,[#]
Department of Computer Engineering, College of Computing and Informatics
University of Sharjah, Sharjah, United Arab Emirates,
Email: anassif@sharjah.ac.ae

Darko Castven[#]
Department of Medicine I, University Medical Center Schleswig-Holstein,
University of Lubeck, Lübeck, Germany
Email: Darko.Castven@uksh.de

Manar Abu Talib
Department of Computer Science, College of Computing and Informatics
University of Sharjah, Sharjah, United Arab Emirates
mtalib@sharjah.ac.ae

Jibran Sualeh Muhammad
Department of Biomedical Sciences, College of Medicine and Health
University of Birmingham, Birmingham, United Kingdom
Email: j.sualehmuhammad@bham.ac.uk

Ahmed Ammar Kubba[^]
Department of Computer Science, College of Computing and Informatics
University of Sharjah, Sharjah, United Arab Emirates
Email: u23103280@sharjah.ac.ae

Jens Marquardt[^]
Department of Medicine I, University Medical Center Schleswig-Holstein,
University of Lubeck, Lübeck, Germany
Email: Jens.Marquardt@uksh.de

Abdalla Sayed Ali
Department of Computer Science, College of Computing and Informatics
University of Sharjah, Sharjah, United Arab Emirates
Email: U22103623@sharjah.ac.ae

*Corresponding Author
#,^: Authors contributed equally to this work

# Abstract

This study explores the use of deep learning and explainable artificial intelligence to diagnose hepatocellular carcinoma (HCC) and define effective biomarkers across five different stages of disease developmental using a transcriptomic biomarker HCC dataset constructed via semi-supervised learning from three source datasets. Several deep learning experiments were conducted with different feature extraction techniques and gene sets to identify the most effective features for training high-accuracy models with minimal loss. The best-performing model, using 15 selected genes with the SelectKBest algorithm, achieved **90.74% accuracy**, while the model with the lowest recorded loss of **0.3187** was obtained using 20 selected genes. To address the issue of class imbalance in the dataset, a weighted training approach was conducted, and for model transparency and interpretability a SHAP-based XAI analysis provided insights into the model's decision-making, consistently finding **DNAJB14** as the most influential gene. Functional validation in this study has provided compelling evidence that DNAJB14 plays an important role in the adverse properties of HCC and that its inhibition effectively reverses tumour cell migration, invasion, colony and sphere formation. The main limitation of this study is the dataset's class imbalance, and while weighted training helped mitigate this, further research and additional data are needed to guarantee model generalizability. Future studies should also explore the influence of genetic variations, environmental factors, and clinical differences on model performance across diverse populations.

# 1. Introduction and Related Work

In the past decade, artificial intelligence and its subsets (deep learning and machine learning) have emerged as revolutionary techniques in addressing the challenges of hepatocellular carcinoma (HCC) by leveraging vast amounts of clinical and diagnostic data, with a high number of studies and researchers exploring the potential of using these techniques for a wide range of medical applications in the context of diagnosis and prognosis of liver cancer patients [1][2][3][4] While data mining and machine learning techniques like association classification have been previously used for medical applications like breast cancer classification [5], the usage of deep learning methods can provide more reliable and accurate models [6], in addition to enabling the usage of more training data through privacy-preservation mechanisms such as federated learning in the context of hospitals and medical data [7][8].

Han et al. [9] have tried to improve early detection of HCC by combining dynamic network biomarker (DNB) analysis with graph convolutional neural networks (GCN) and with the use of multi-omics, which are used to study HCC development in mice using transcriptomic data and gene expression levels, using a five-year relative survival rate of HCC patients dataset from the NCBI with addition to RNA-seq data uploaded to the NCBI database (multi-omics gene expressions data) [10]. The DNB model identified the critical transition point at 7 weeks of age, achieving a 100% accuracy in classifying healthy and cancerous mice while also accurately predicting the health status of newly introduced mice. However, our approach was different by using feature selection to detect the biomarkers for detection of HCC.

Purba et al. [11] used Deep Neural Networks to classify cancer, with training using MicroRNA data analysis. Three types of data normalization (Min-Max, Sigmoid, Softmax) and three activation functions (ReLU, Sigmoid, TanH) were compared. The best accuracy (98.33%) out of four experiments (with no feature selection) in classifying HCC with MicroRNA data was achieved using Min-Max data normalization, ReLU activation function, and batch normalization with specific parameters. The microRNA dataset was obtained from the GDC Data Portal of the National Cancer Institute [12]. The data used in this study consisted of 600 data divided into 300 data on HCC class and 300 data on normal class with a total of 328 MicroRNA features.

Zhang et al. [13] aimed to create a patient similarity network using three types of HCC omics data and introduced a novel diagnosis method. This method combined similarity network fusion, denoising autoencoder, and dense graph convolutional neural network to use patient similarity networks and multi-omics data (DenseGCN). Comparisons with other machine learning methods on the TCGA-LIHC dataset demonstrated that the proposed approach outperforms them in all metrics. The proposed method achieved an accuracy of up to 0.985, using Liver Hepatocellular Carcinoma (LIHC) omics dataset. Like our approach, they have used feature selection techniques, but they used a denoising autoencoder.

Ahn et al. [14] review recent studies that have applied deep learning in risk prediction, diagnosis, prognosis assessment, and treatment planning for HCC patients. It also discussed different models of deep learning like ResNet, Auto Encoders, and CNNs, using several datasets like the TCGA, VHA, and Taiwanese NHIRD. Ioannou et al. [15] used an RNN to predict HCC development within three years, getting superior performance with an AUC of 0.759. Phan et al. [16] applied a CNN to predict HCC among patients with viral hepatitis, achieving an AUC of 0.886 and 0.980 accuracy. Nam et al. [17] constructed DNNs for predicting HCC incidence and recurrence after liver transplantation, outperforming traditional models with higher C-index values. As shown, the highest accuracy was 0.980. According to the study, there were certain limitations; Serum AFP, a predictive biomarker for hepatocellular carcinoma [18][19][20], has limited sensitivity for early detection. Multi-omics research is needed for better biomarkers. Deep learning algorithms generalizability and lack of public availability hinder external validation for reliability and effectiveness.

Das et al. [21] aimed to build a new technique for detecting cancer by a new system called watershed Gaussian based deep learning using the GMM algorithm and built a DNN model using a CAD model and WGDL technique for classification with a 99.38% classification accuracy and a validation loss of 0.062. They used CT images as a model parameter and a dataset consisting of 225 CT images of HCC patients from IMS and SUM hospitals in India, from 75 different HCC and metastatic carcinoma patients. A level co-occurrence matrix (GLCM) method was used for classification and feature selection. One limitation is they developed a DNN using Keras that uses Tenserflow for their feature selection to classify the disease accurately, it required to balance the data set otherwise the classifier will produce an error function. A GE medical System CT scan machine for recording the images was used as a clinical test.

Tao et al. [22] demonstrated the potential of recognition for early stages cancer detection using a whole-genome (SCNA) profiling method with also the use of machine learning by also taking advantage of a method called "Liquid biopsy" as the detection of early stages cancer are more challenging as the tumour cells that carry the cancer genomic aberrations tend to release much less DNA into the blood, which would make it impossible to diagnose because we will need an unrealistic amount of blood from the patient, and that where comes the use of the machine learning algorithms used in this research such that wRF-driver algorithm can take advantage of the SCNA profiles and also some external provided medical data (TCGA – ICGC). The ML model was built using a discovery cohort of 209 patients, with an overall area under curve (a commonly used metric in machine learning to evaluate the performance of a binary classification model) of 0.893 / 0.874 for early-stage, and 0.933 for patients with only Stage 0 - A HCC over the two cohorts, The accuracy in validation cohort 2 is noticeably lower than validation in cohort 1, likely due to more patients with very early stage tumour in validation cohort 2.

Książek et al [23] conducted a new approach for detecting HCC patients using 10 ML algorithms and a dataset of 165 HCC survival patients from CHUH in Portugal and results were noted by RF algorithm F1-Scores and did 2 different experiments. RF algorithm conducted the highest F1 score of 1 out of 1 and for the linear SVM highest score was 0.7804 during testing. One limitation for this design was the lack of features, which we aimed to overcome in our project. Different experiments reviewed by Chen et al. [24] have shown the role of biomarkers in HCC detection using RNAs and identifying tumour antigens to activate anti-tumour immune responses, while also using blood biopsy as a clinical test. Petinrin et al [25] review existing studies on classical machine learning and deep learning models in metastatic cancer, it demonstrates the use of deep learning for image data extraction, noting that feature extraction capabilities of deep learning aid the development of better and more accurate models, hence why we use feature selection in this experiment.

Jiang et al. [26] demonstrates the importance of histopathology image analysis in HCC diagnosis and prognosis, and how deep learning has become a leading technique in this domain for classifying and localizing HCC and tumour cells, it surveys over 50 studies in this field, the main methodologies used were supervised learning, weak supervised learning and multi-scale and multi-resolution networks. Amongst the studies overviewed there were different public datasets used like the TCGA-LIHC tissue slide images and related clinical information dataset, PAIP 2019 dataset for segmentation tasks from MICCAI 2019 Grand Challenge, and the KMC dataset providing diverse histopathological images for different HCC subtypes. Limitations were mostly data and image related, like colour inconsistencies and large image sizes, and many DL models are considered "black boxes," meaning their decision-making processes are not transparent or easily interpretable by clinicians. For performance metrics some accuracies were reported as high as 0.98 in some studies.

Ngan et al. [27] showed a review of how machine learning is used to analyse mass spectrometry-based metabolomics data for early HCC diagnosis. This technique helps to capture metabolic changes associated with cancer and enables screening out physiological biomarkers of cancer risk and clinical biomarkers of cancer which can help in identifying it in early stages. Methodologies used were mass spectrometry, machine learning with different models like Random Forests (RF), Principal Component Analysis (PCA), Support Vector Machines (SVM), Partial Least Squares-Discriminant Analysis (PLS-DA) and Neural Networks (NN), with the help of a feature selection technique called correlation-based feature selection (CFS) with logistic regression and Lasso regression. Certain limitations were noted like the exponentially increased data volume and complexity of the mass spectrometry data and the lack of clinical validation. For liver cancer detection, supervised ML models such as linear SVM and logistic regression demonstrated high accuracy (>85%) for different studies on tissue/serum.

Matboli et al. [28] aimed to leverage machine learning techniques to diagnose early HCC using RNA signatures with laboratory parameters. It uses a comprehensive dataset which contains RNA expression levels and clinical parameters from 267 subjects including 102 malignant HCC, 67 of which are benign liver conditions and 98 healthy controls, in addition to access of the GSE14520 dataset. The RNA signatures include mRNAs (RAB11A, STAT1, ATG12), miRNAs (miR-1262, miR-1298, miR-106b-3p), and lncRNAs (RP11-513I15.6, WRAP53) select on their known involvement towards HCC pathogenesis (in key biological pathways). Different machine learning models were evaluated like KNN, RF, SVM, LGBM, DNN trained and tested using a 70/30 dataset split. LGBM was the best performing model, achieving an accuracy of 98.75% superior to other classifiers. The study also employed feature selection techniques. The model also included 22 features (age, sex, smoking, cirrhosis, non-cirrhosis, albumin, ALT, AST bilirubin (total and direct), INR, AFP, HBV Ag, HCV Abs, RQmiR-1298, RQmiR-1262, RQmiR-106b-3p, RQmRNARAB11A, and RQSTAT1, RQmRNAATG12, RQLnc-WRAP53, RQLncRNA- RP11-513I15.6). Future work includes further validation and testing of the model, exploring the model's potential in also predicting disease progression and response to treatment. Limitations were small and unreliable data sizes.

Shen et al. [29] aimed to construct risk prognosis model by integrating multiple key genes related to aging and to explore the relationship between risk score and immune cell and tumour microenvironment by combining bulk and single-cell sequencing using a NMF machine learning model where clinical data and survival status acts as model parameters, using a dataset from TCGA (374 cancer samples and 50 normal samples of clinical data). Without feature selection, for the TCGA training set, the model achieved 0.797, 0.749, 0.740 for 1, 2, 3 years amongst several experiments.

Gholizadeh et al. [30] aimed to identify key mRNAs which can serve as biomarkers for diagnosing HCC, it uses a non-fusion integrative multi-platform meta-analysis to integrate gene expression data from different platforms, then applies machine learning methods to develop diagnostic and prognostic models. Data is used from many gene expression platforms like Illumina and Affymetrix datasets, with a total of 939 samples, comprising 493 tumour and 446 non-tumour samples, were used. The datasets include GSE57957, GSE39791, GSE36376, GSE84005, GSE12941, GSE64041, GSE45267, and GSE84402, sourced from the Gene Expression Omnibus (GEO) and The Cancer Genome Atlas (TCGA). The study utilizes the Linear Models for Microarray Data (LIMMA) R package to identify differentially expressed genes (DEGs). A Bayesian approach is employed to assess statistical significance, with a p-value threshold of 0.05 and a fold-change cutoff of 1. One limitation noted could be potential biases due to batch effects despite correction efforts, and for future work they suggested enhancing the robustness of the identified biomarkers and exploring additional molecular signatures to improve diagnostic and prognostic accuracy.

*Table 1 Overview of the existing literature in terms of datasets, objectives, methodology, and results compared to our work*

| Ref | Datatype | Task | Dataset | Feature selection | Performance | Clinical test |
|---|---|---|---|---|---|---|
| [31] | Multi-omics gene expressions | HCC Detection | NCBI | × | 100% accuracy | × |
| [32] | Multi-omics gene expressions | HCC Detection | NCBI | × | 100% accuracy | × |

| [33] | DNA sequences | HCC Detection | NCBI | ✗ | CNN Model: 80.36% accuracy vGG16 model: 98.86% accuracy fine-tuned vGG16 model: 100% accuracy | ✗ |
|---|---|---|---|---|---|---|
| [34] | microRNA expressions | HCC Detection | GDC Data Portal NCI 2018 | ✗ | Best accuracy is 98.33% | ✗ |
| [35] | LIHC omics data | HCC Detection | LIHC omics datasets from TCGA | Denoising autoencoder | 0.9857 Accuracy | ✗ |
| [36] | CT images | HCC Detection | Datasets from IMS and SUM hospital (India) | GLCM | 99.38% classification accuracy | GE medical system scan machines |
| [37] | SCNA genome profiles & CNA plasmas | HCC Detection | Discovery cohort of 209 HBV patient's genome-wide SCNA profiles in early stages | Gini-Impurity Index | AUC of 0.920 for cohort 1, AUC of 0.812 for cohort 2 | AFP Screening |
| [38] | 23 quantitative features and 26 qualitative features of HCC patients' medical information | HCC Detection | CHUC | Genetic algorithm with Cross validation | f1 score of 1 out of 1 and for the linear SVM highest score was 0.7804 during testing | ✗ |
| [2] | histopathology images / tissue slides images and related clinical information | HCC Diagnosis and treatment | TCGA-LIHC tissue images, PAIP 2019, KMC | CNN models was used | Varies, reaching as high as 0.98 in some experiments | ✗ |
| [39] | liver tissue, serum samples and other clinical info | Detection of different types of cancer | Private dataset of liver tissue and serum samples and including data from another studies | CFS, Logisitic Regression, Lasso Regression | For the liver cancer experiments, good accuracies are reported, 80% and above for tissue and serum studies, however there is no single accuracy denoted. | ✗ |
| [40] | RNA expression levels and clinical information,Biochemical markers | HCC Diagnosis | A private dataset with clinical and RNA expression data from 267 subjects, including 102 with malignant HCC, 67 with benign liver conditions, and 98 healthy controls. Additionally, the GSE14520 dataset was used | ✗ | LGBM was the best performing model, achieving the highest accuracy of 98.75% | ✗ |
| [41] | 374 cancer samples and 50 normal samples clinical data. | predict prognosis and immunotherapy efficacy | TCGA | ✗ | For TCGA training set: AUC values of 0.797,0.749,0.740 for 1,2,3 years. | ✗ |
| [42] | gene expressions | identify key mRNAs which can server as biomarkers for diagnosing HCC | GSE57957, GSE39791, GSE36376, GSE84005, GSE12941, GSE64041, GSE45267, GSE84402, TCGA and GEO | ✗ | p-value threshold of 0.05 and a fold-change cutoff of 1 | Serum Biomarker Analysis, Liver Function Tests, Reverse Transcriptio |
| | | | | | | n and qRT-PCR, Diagnostic Imaging and Pathology |
| **Our Work** | Multi-omics gene expressions | Multi-Stage HCC detection and identification of relevant biomarkers for early-stage HCC | Private Lubeck data, GSE89377 , TCGA | SelectK Best, Ant colony, Genetic Algorith m, Particle Swam Optimiz ation | 90.74% best accuracy and 0.3187 loss | Functional in vitro target evaluation |

As observed in this literature review, studies tend to explore different biomarkers, including mRNAs and miRNAs, to differentiate between healthy and cancerous patient tissue and different HCC stages. However, many drawbacks to the reviewed papers can be noted, which include limited dataset sizes, the need for better feature selection approaches, and a lack of clinical validation in addition to the 'black box' nature of deep learning making interpretability difficult and raising further issues with applying machine learning to medical areas. This study further explores the potential of using genomic /RNA expression data for training deep learning models for multi-stage HCC classification. Our contributions to the existing research are as follows:

- Our deep learning model can classify patient tissue samples into hepatocarcinogenesis including non-malignant and malignant tissue, whereas machine learning models used in other studies often have less detailed or broader classification capabilities that predict a smaller number of HCC classes.
- The HCC dataset used for training the deep learning model was constructed from one private dataset and two public datasets using semi-supervised learning and contains five categories, with 770 patient samples represented by 11,150 gene expression features.
- The final accuracy of the deep learning model after performing the necessary tests was recorded at 59% on the initial set of genes, 61% on the second set of genes, and 86% on the third set of genes, each extracted from one of the three datasets during the feature extraction phase. Additionally, the model performed best with the SelectKBest algorithm with a classification accuracy of 90.74% at 15 selected genes.
- Our study implements an explainable AI approach to contribute to overcoming the black box issue noted by previous studies using AI models for medical applications, increasing the reliability and transparency of our model as a result, in addition to bringing attention to a gene that can potentially have links to HCC that have not yet been established in the literature (DNAJB14).
- Our study performs in vitro functional validation on the DNAJB14 gene to validate links to HCC.

This paper is organized as follows: The first section introduces the research topic (deep learning and HCC), along with background information and a review of the related literature, noting existing limitations in the literature. The second section outlines the research methodology, detailing dataset preprocessing, feature extraction, the experimental setup, and the explainable AI approach. The third section presents and describes the results of the experiments. The last section concludes the paper and presents the study's limitations, potential future work.

# 2. Research Methodology

This section describes the methodology of the project in detail, as observed in Figure 1, which consists of data pre-processing on the HCC dataset which consists of three source datasets and feature extraction on the Lubeck, TCGA, and HCC dataset. Afterwards, the deep learning model is trained on the HCC dataset using the extracted gene sets as input features, and then the models are evaluated accordingly. Finally, the explainable AI approach is implemented, and the results are interpreted in the last section.

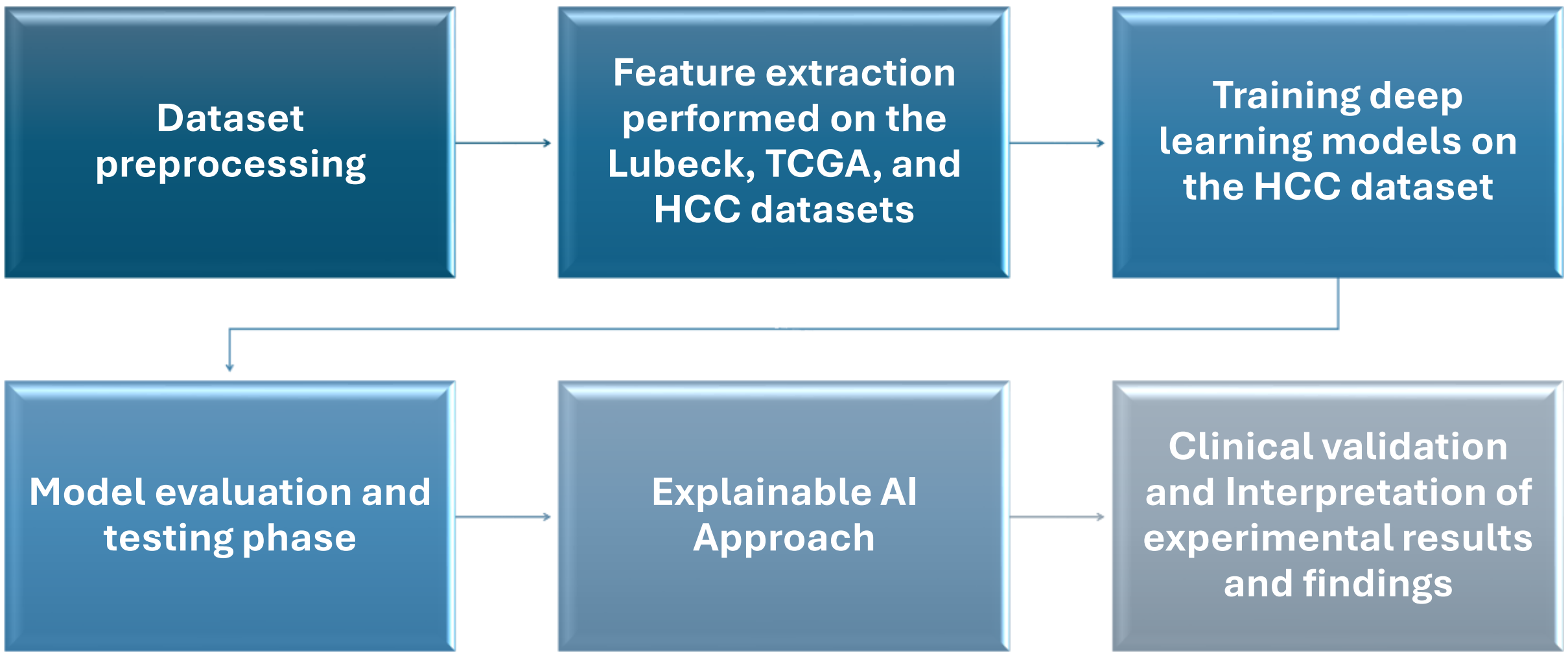

*Figure 1 Paper methodology summary*

## 2.1 Dataset Preprocessing and Feature Extraction

The training dataset is a specialized hepatocellular carcinoma dataset containing 770 patient samples labelled in five categories and represented by genomic expression data features. Each samples contains 11,150 corresponding genomic expression levels, which represent the input training features for the model to predict the developmental HCC stage of the target patient. This dataset was constructed using semi-supervised learning from three source datasets: the private "Lubeck" dataset from medical domain experts, and the two publicly available datasets, TCGA [43] and GSE89377 [44]. As can be observed in Figure 2, which visualizes the distribution of the dataset's classes, the most common samples in the dataset belong to the 'phcc' and 'ehcc' classes, whereas the 'sl' class has the fewest samples at less than 50 instances. This class imbalance in the dataset is accounted for in the experimental setup section by utilizing different weights for the classes during training. The dataset's patient tissue samples are classified into five categories:

1. **Surrounding Liver (sl):** which is absent of any HCC gene expression.
2. **Early Hepatocellular Carcinoma (ehcc):** representing an early-stage HCC diagnosis.
3. **Progressed Hepatocellular Carcinoma (phcc):** indicating an advanced stage of HCC.
4. **Low-Grade Dysplastic Nodules (lgdn):** which are associated with a lower risk of HCC development.
5. **High-Grade Dysplastic Nodules (hgdn):** which are linked to a higher risk of HCC progression.

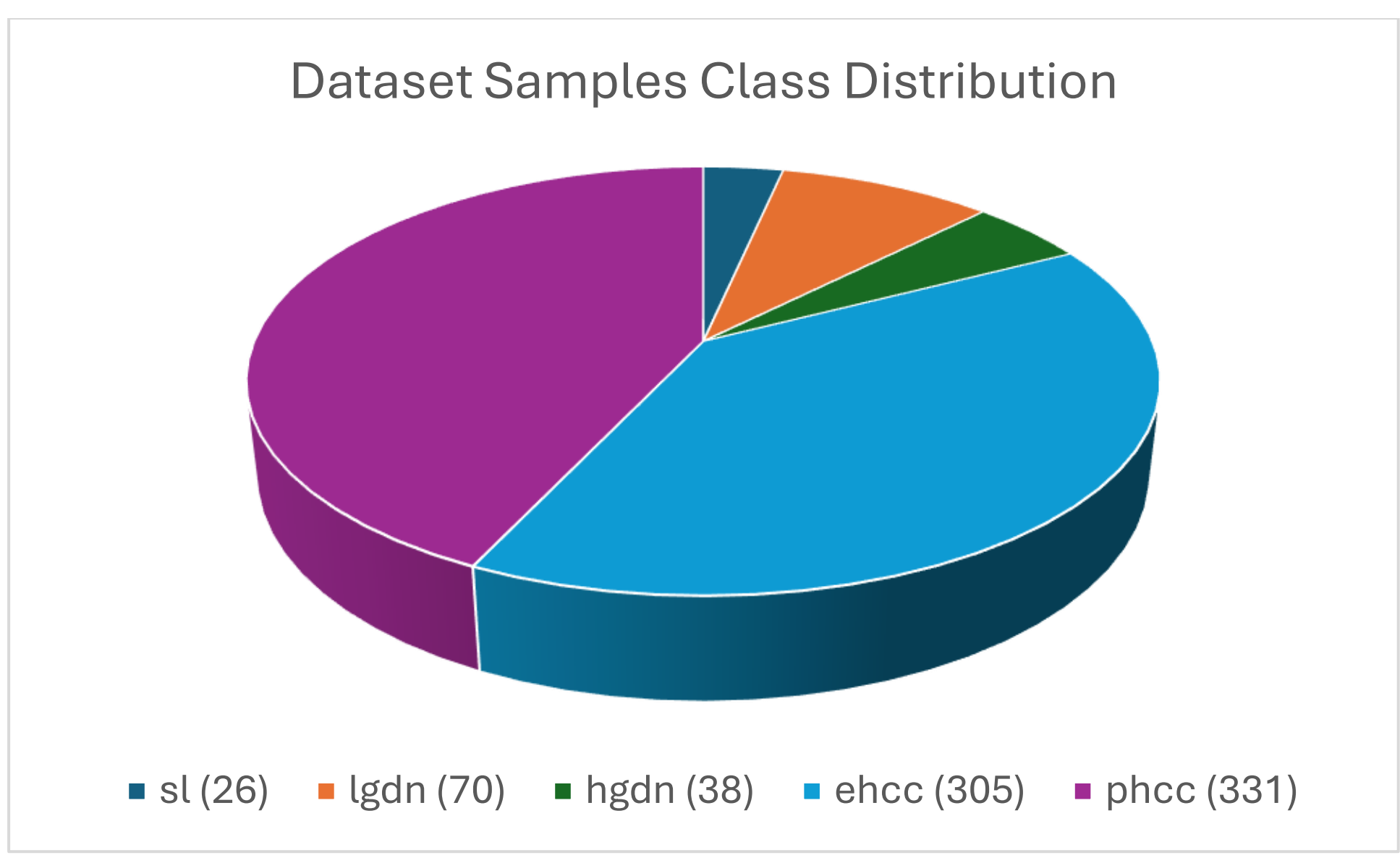

*Figure 2 Visual representation of the dataset class distribution.*

The deep learning model was trained in three separate experiments using three different sets of genes as observed in Table 1, which were extracted using feature selection that was performed on the Lubeck and TCGA source datasets, in addition to the HCC dataset which was constructed using semi-supervised learning. Conducting these three experiments enables us to compare the performance of the model on each selected gene set for diagnosis, minimizing possible bias in the results and providing us with more insights into the effectiveness of using different genomic markers as input features, improving the model's robustness and its applicability in clinical settings in addition to providing future researchers with insights into possibly significant genetic biomarkers.

*Table 2 Feature extraction gene sets used in the model's training*

| **Gene Set** | **Selected Genes** | **Source Dataset** |
|---|---|---|
| **Set 1** | **['SELENBP1', 'MASP1', 'PPP1R14B', 'TAF2', 'OTUB1']** | **Lubeck** |
| **Set 2** | **['HGF', 'PDGFRA', 'HIF1AN', 'SAE1']** | **TCGA** |
| **Set 3** | **['IFNA5', 'IVD', 'DNAJB14', 'ITGB8', 'PADI4']** | **HCC** |

## 2.2 Experimental Setup

The deep learning model used in this project is a feedforward neural network which was built using the Keras python API for machine learning. The model, as visualized in Figure 3, consists of an initial input layer whose shape matches that of the training data, to enable feeding the training data into the neural network. Afterwards, there are two hidden dense layers which serve their own purposes. The first dense layer consists of 64 neurons and a Rectified Linear Unit (ReLU) activation function which adds non-linearity to the model, an important characteristic for learning complex data patterns in machine learning.

After the first dense layer, a dropout layer is introduced with a 20% dropout rate. This means that this layer randomly zeroes 20% of the input data during each training round, which is important for preventing overfitting as it ensures that the DL model does not overtly rely on any specific set of neurons. The second dense layer is identical to the first alongside its dropout layer except for the fact that it uses 32 neurons instead. The final layer in the model is a dense layer that uses five neurons, equivalent to the five classes found in the HCC dataset. This layer uses a linear activation function, which is needed for the loss function that follows it.

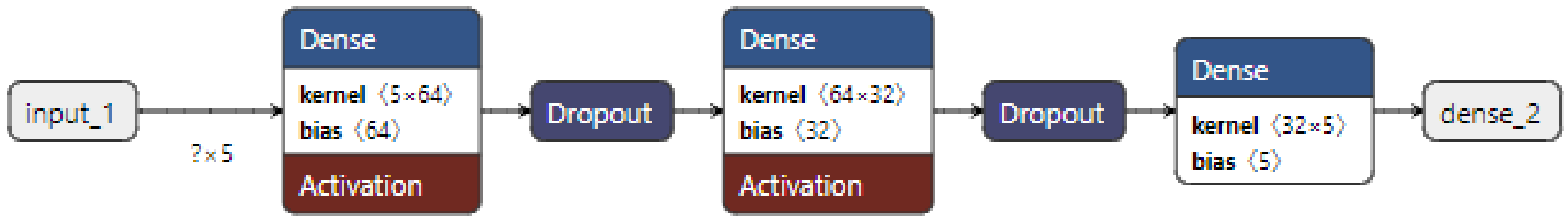

*Figure 3 Deep learning model visualization*

The deep learning model was compiled using the Adam optimizer and the sparse categorical cross entropy loss function, which is the appropriate loss function for multi-class classification machine learning tasks in which the output layer doesn't use a softmax activation function. Additionally, the model utilized a learning rate scheduler which initially maintains a high learning so that the model can learn from the training data quickly. After 40 epochs, the LR scheduler reduces the learning rate and then further reduces it after 80 training epochs. This enables refining the learning process and allowing the model to make very precise updates throughout the training process, leading to overall better model stability during training. The model also utilized an early stopping mechanism which stops the training process when the model is no longer improving on the validation set. This approach can be useful for preventing overfitting and avoiding a lengthy training process with little to no benefits.

The training set consisted of 35% of the original dataset, whereas the test and validation splits were 48.75% and 16.25% of the dataset, respectively. The model's training was conducted on a workstation using the Windows 10 operating system with 32 gigabytes of RAM and an Nvidia RTX 4000 GPU. The CPU of the workstation, which was utilized during the model's training, is Intel® Xeon® W-2102 with four cores and a base speed of 2.90 GHz.

Because of the relative imbalance of classes in the training dataset, which can be observed in Figure 2, we have utilized a weighted training approach to minimize training bias in the DL model. The weighted training approach is enabled by fitting the model with a dictionary containing a different weight for each class in the dataset, with a total of five weights for five classes. These weights are calculated using a special computing function from the scikit-learn library which produces the weights based on the relative frequency of each class. As can be observed in Table 3, the assigned weight for each class is inversely proportional to the relative frequency of that class, i.e. the more frequently a particular class occurs in the dataset samples, the smaller the assigned weight to that class will be to indicate to the model that it should prioritize the least frequent minority classes over the majority classes during the training process.

*Table 3 Assigned weights for each class in the dataset*

| Class Name | Frequency | Proportion | Assigned Weight |
|---|---|---|---|
| **ehcc** | 305 | 39.61% | 0.505 |
| **hgdn** | 38 | 4.935% | 4.053 |
| **lgdn** | 70 | 9.091% | 2.200 |
| **phcc** | 331 | 42.99% | 0.465 |
| **sl** | 26 | 3.377% | 5.923 |

## 2.3 Explainable AI Approach

As transparency is highly valued when it comes to the integration of AI in the medical field, an explainable AI approach is also explored for the purpose of producing a medically trustworthy and transparent solution. Three explainable machine learning models, consisting of two Random Forest classifiers and one XGBoost classifier with each using different optimization algorithms, are trained on the HCC dataset and then explained afterwards using the SHAP (SHapley Additive exPlanations) library, which helps explain the predictions of the models and their contributing features using a game theory approach [45][46]. The explainable AI approach of our study is summarized in Figure 4.

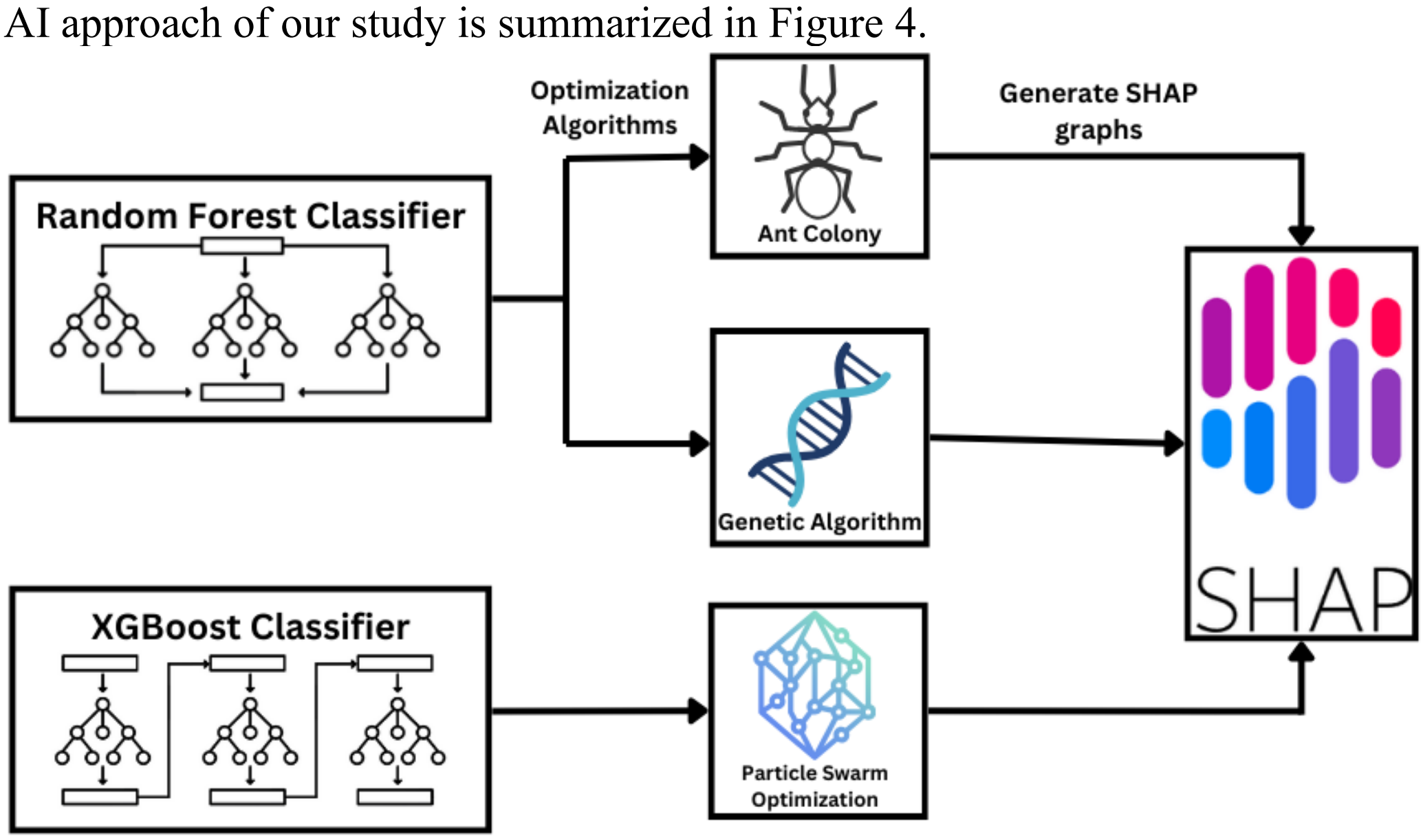

*Figure 4 Explainable AI Methodology*

The first two models consist of a random forest classifier, with the first model being trained using the Ant Colony optimization algorithm for feature selection, and the second model using the Genetic Algorithm. The third model consists of an XGBoost classifier that was trained using the Particle Swarm optimization algorithm. SHAP is used to explain all three models, and the results are cross verified between the models to minimize possible model bias in the results by not relying on a single model or optimization algorithm.

The first two models consist of a Random Forest classifier, with the first model being trained using the Ant Colony optimization algorithm for feature selection, and the second model using the Genetic Algorithm. Random Forest is a type of ensemble learning technique which builds multiple decision trees during the training process and then averages their total outputs during the prediction process [47]. It is based on a technique called Bagging (or Bootstrap Aggregating), in which different subsets of the dataset are randomly sampled with replacement to train individual decision trees [48]. This process reduces variance and enhances model stability, making it more robust against overfitting, especially when working with high-dimensional medical datasets like our HCC dataset.

The third model consists of an XGBoost classifier that was trained using the Particle Swarm optimization algorithm. XGBoost (Extreme Gradient Boosting) is a highly efficient gradient-boosting model which builds decision trees sequentially, with each new tree improving the accuracy of the trees preceding it [49]. This technique is based on Boosting, another ensemble learning approach, which reduces model bias by iteratively adjusting the model weights of a weak learner based on misclassified samples in each iteration of the model, leading to a strong final predictive model [50].

Ant Colony Optimization is inspired by the foraging behaviour of ants, where they deposit pheromones to mark the best paths to food sources. ACO simulates this behaviour by iteratively building feature subsets and prioritizing the most promising features based on a chosen performance metric, improving the classification performance of Random Forest while maintaining its interpretability [51].

The Genetic Algorithm is an optimization algorithm inspired by natural selection and genetics. It keeps a population of potential feature subsets, which evolve over multiple iterations using selection, crossover, and mutation techniques. The best feature sets are chosen based on a predefined fitness function to maximize classification performance [52].

Particle Swarm Optimization is an algorithm inspired by collective bird movement, in which each potential solution moves based on its own experience and the best solutions found by other particles. When used for of feature selection for XGBoost, PSO adjusts the weights of different features iteratively, improving the model's ability to generalize with each iteration and eventually leading to an optimized feature subset with minimal redundancy [53].

SHAP is used to explain all three models, and the results are cross verified between the models to minimize possible model bias in the results by not relying on a single model or optimization algorithm. By integrating both Bagging-based (Random Forest) and Boosting-based (XGBoost) models with several iterative optimization algorithms, this approach ensures a comprehensive analysis that balances accuracy, interpretability, and robustness in medical AI applications.

## 2.4 Clinical Validation Methodology

### 2.4.1 Survival analysis using public dataset

Gene Expression Profiling Interactive Analysis (GEPIA2) online tool (http://gepia2.cancer-pku.cn/#index) was used to evaluate the influence of DNAJB14 on overall survival in HCC patients retrieved from the publicly available knowledge of The Cancer Genome Atlas Program (TCGA) database. Cutoff values for high and low expression were set to 80%.

### 2.4.2 Cell culture

The Huh7 cell line was obtained from Riken Cell Bank. Cells were grown in Dulbecco's modified Eagle's medium (DMEM), supplemented with 2 mM L-glutamine, 1 unit/mL penicillin/streptomycin, and 5% fetal bovine serum (FBS). Cell line was maintained at 37°C and 5% CO2, routinely tested and authenticated by STR testing.

### 2.4.3 Nucleic acid extraction

Total RNA was extracted using Monarch Total RNA Miniprep Kit (NEW ENGLAND Biolabs) according to the manufacturer's instructions. RNA quality and purity was determined by Nanodrop One spectrophotometer (Thermo Fisher Scientific).

### 2.4.4 Quantitative real-time polymerase chain reaction

Two-step reverse-transcription quantitative polymerase chain reaction using iScript cDNA Synthesis kit (Bio-Rad Laboratories), SYBR Green Master-Mix (Bio-Rad Laboratories) and CFX Connect (Bio-Rad Laboratories) was performed. Oligonucleotide primers were designed using Primer3 v.0.4.0 (https://bioinfo.ut.ee/primer3-0.4.0/). Forward: 5' TTC ATC ACC AGC ATC TTT CG 3' Reverse: 5' CAG GTA GTG GCG ATC AAA GC 3' (Eurofins).

### 2.4.5 Western blotting

Cell lysates were prepared using M-PER tissue extraction buffer (Thermo Fisher Scientific) containing Halt Protease & Phosphatase Inhibitor Cocktail (Thermo Fisher Scientific). 20 µg of protein lysate were separated by SDS-PAGE and transferred onto nitrocellulose membrane (GE Healthcare Life Sciences). Membranes were probed with the DNAJB14 antibody (rabbit, polyclonal, proteintech #16501-1-AP).

### 2.4.6 siRNA-mediated knockdown

For transfection, cells were seeded at low confluency (30-50%) in 6-well plates. After 24 h 50, 100 and 150 pmol siRNA (non-sense control and siRNAs targeting DNAJB14, ambion) was introduced by using transfection reagent Lipofectamine 2000 (Thermo Fisher Scientific) following the manufacturer's instructions. Subsequent functional assays were conducted 72 h after transfection.

### 2.4.7 Viability assay

Cell viability was determined using a colorimetric assay (WST-1) following the manufacturer's protocol (Roche). In total, $5\times10^3$ cells were plated on 96-well plates; after overnight incubation, cells were transfected with increasing concentrations of DNAJB14 siRNA reflecting concentrations of 50, 100 and 150 pmol siRNA from 6-well plate experiments and the viability was measured after 72 hours.

### 2.4.8 Colony and sphere formation assays

Cells were treated for 72 h with siRNA-mediated knockdown of DNAJB14. $1x10^3$ cells were plated on 6-well plates for colony formation (CFU) and $1x10^3$ cells were plated on 48-well plates for sphere formation (SFU) assay into a semisolid soft agar (Carl Roth). After 14 days colony and sphere formation capacities were determined and represented as number of colonies/spheres for each treatment group in comparison to the control group.

### 2.4.9 Migration assay

After siRNA-mediated knockdown of DNAJB14, migration was evaluated with Culture-Inserts in a µ-Dish 35mm (Ibidi) and compared with control cells. To distinguish between migration and proliferation, cell proliferation was suppressed by using the proliferation inhibitor Cycloheximide (2 µM, Carl Roth). In total $5x10^4$ cells were seeded on Culture-Inserts in DMEM (10% FBS). The cells were allowed to attach for 24 h before the culture inserts were removed. Cycloheximide was administered 6 h prior to treatment. After incubation, the culture inserts were removed and the plate were filled with 2 ml culture medium containing and the cells were monitored immediately after gap formation and images were taken at 0, 24, 48, 72, 96 and 120 h using a phase contrast microscope.

### 2.4.10 Invasion assay

Cell invasion was determined using the xCELLigence DP system (Roche, Mannheim, Germany). For the invasion assay, a total of 80,000 Huh7 cells with siRNA-mediated knockdown of DNAJB14 or control cells were seeded on CIM-Plates 16. To assemble the CIM-Plates 16, the bottom of the upper compartment was coated with 30 µl of collagen I (1 mg/ml) and IV (50 µg/ml) and allowed to dry. Cells were applied to the upper chamber of the CIM-Plate, which was then placed in the RTCA DP instrument and connected to the computer. The real-time cell invasion assay was monitored for the next 72 h with data acquisition at 15' intervals. Monitoring and analysis were performed using xCELLigence RTCA software (version 1.2.1., Roche, Germany).

### 2.4.11 Statistical analyses

Statistical analyses were performed using one-way ANOVA. $p \leq 0.05$ were considered statistically significant. All results were presented as means ± standard deviation (SD) from at least three independent experiments.

# 3. Results and Discussion

This section details the results of the deep learning model and its performance, in addition to the explainable AI approach and the most significant genes based on the conducted SHAP analysis. With informed advice and insights from medical domain experts, a thorough discussion and interpretation of the results and findings is also presented.

## 3.1 Deep Learning Model Performance

The deep learning model was evaluated on the test split, which represents 35% of the original dataset. As the training process consisted of three separate experiments, in which the model was trained on each selected genes set, the model was evaluated after each experiment to determine the set of genes which produced the best-performing DL model.

| Training Gene Set | Accuracy | Loss |
|---|---|---|
| **Set 1** | **0.5926** | **0.9591** |
| **Set 2** | **0.6148** | **0.9029** |
| **Set 3** | **0.8667** | **0.3531** |

*Figure 5 Model testing results based on each separate gene set*

As can be observed in Figure 5, the results for Set 1 show an accuracy of 59.26% and a loss of 0.9591. This indicates that the model correctly classified 59.26% of the training samples after being trained using this specific gene set. The high loss, especially relative to the loss value in the other experiments, shows that the model is struggling to fit the training data correctly due to high errors in its predictions. The relatively low accuracy and high loss values imply that the model is underfitting, which means that it has failed to learn the data patterns sufficiently which can be due to several reasons, including a poor choice of selected genes, insufficient model layer complexity, or a lack of sufficient training data. As can be observed for the next experiments, the performance metrics improve with different selected gene sets, which strongly indicates that the first reason is the best explanation for the current results.

For Set 2, the model achieves a slightly higher accuracy of 61.48% and a slightly lower loss value of 0.9029 compared to Set 1. This improvement in both accuracy and loss values suggests that the model is performing better using this gene set for training. The decrease in loss means that the model's predictions are closer to the actual values than the previous model, although the accuracy value remains unsatisfactory at only 61.48%. This result points to a slightly better fit to the training data, but there is still room for improvement either in the model architecture, the training data, or the gene selection. As can be observed in the next experiment, improving the gene selection produces the biggest difference in terms of model performance.

Set 3 demonstrates the most significant improvement in performance relative to the previous experiments, with a much higher accuracy of 86.67% and a much lower loss of 0.3531. The high accuracy indicates that the model correctly classified most of the training samples, while the low loss means that the model's predictions are very close to the actual values. This big improvement can be clearly attributed to the selected genes set containing more informative and useful features for training the model, allowing it to learn the underlying patterns in the data accurately and producing better accuracy and loss values as a result.

*Table 4 SelectKBest Experiments testing different numbers of 'K' best genes selected.*

| ‘K’ Best Genes Selected | Accuracy | Loss |
|---|---|---|
| **5** | 87.41% | 0.4372 |
| **10** | 90.37% | 0.3729 |
| **15** | 90.74% | 0.3625 |
| **20** | 90.37% | 0.3187 |
| **25** | 88.89% | 0.3638 |

Additional experiments were carried out on more gene sets which were obtained using the SelectKBest feature selection approach on the HCC dataset, which chooses the top K most relevant features based on statistical tests to reduce data dimensionality and noise [54]. SelectKBest has also been used in a medical context in the existing literature [55]. The experiments ranged from 5 best genes selected to the top 25 best genes. As can be observed in Table 3, the best model performance was recorded at 15 ‘K’ best genes selected, with the model performance in terms of accuracy and loss deteriorating with more genes than 15 in subsequent experiments, except for model loss marginally improving at 20 selected genes.

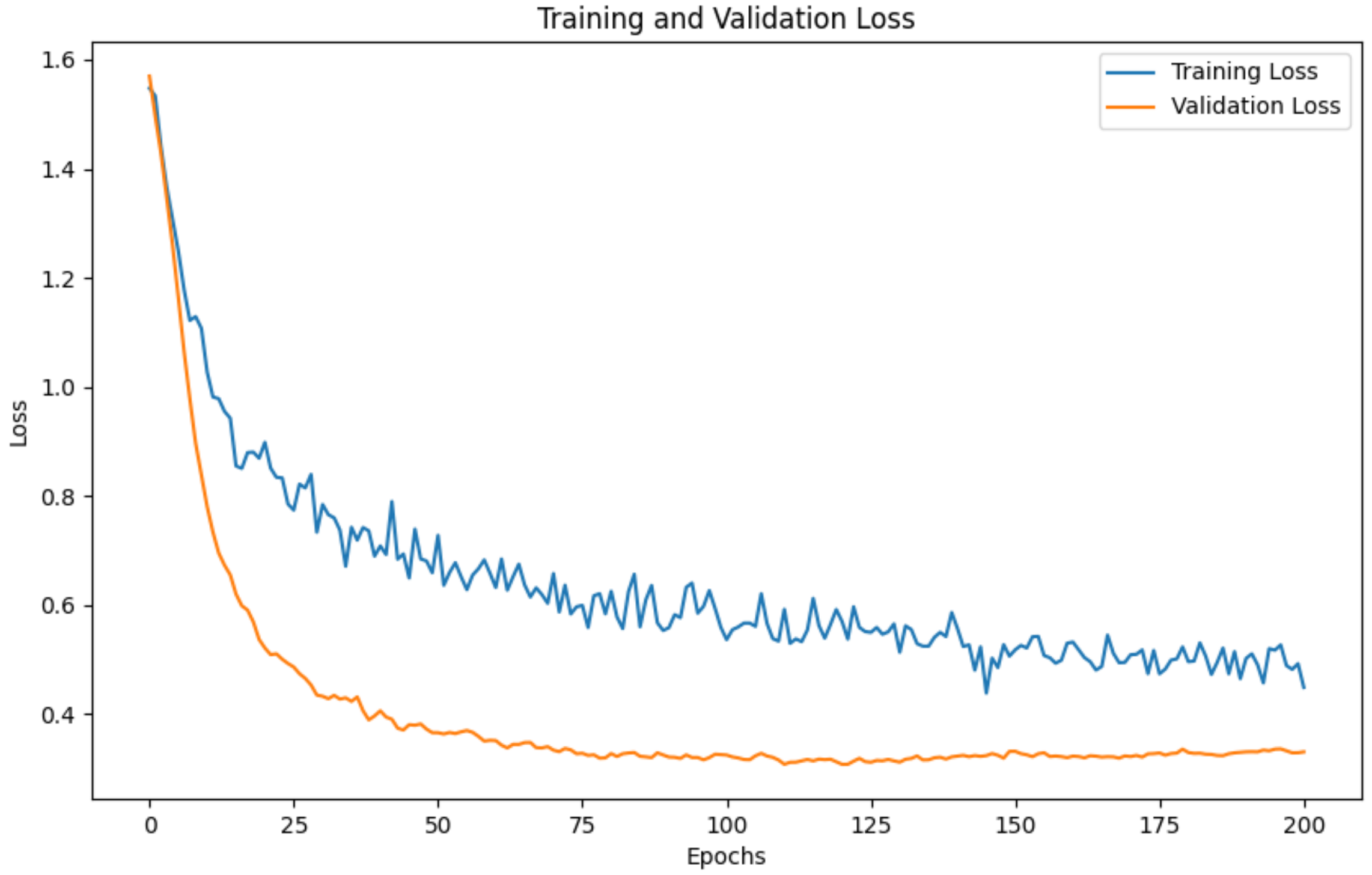

*Figure 6 Training and validation loss graph for the Set 3 training experiment*

The training and validation loss throughout the training process of the Set 3 model can be observed in Figure 6, which shows a smooth progression of validation loss decreasing throughout the training epochs, while the training loss suffers from some fluctuations, it still follows an overall trend of decreasing as the training process continues. Figure 7 displays the training and validation accuracy throughout the training experiment. As can be observed, the training and validation accuracy closely follow the same trend of small increases with some fluctuation during the training process. Overall, both graphs indicate no underfitting or overfitting in the model during the training process and high-performance metrics by the end of the model's training.

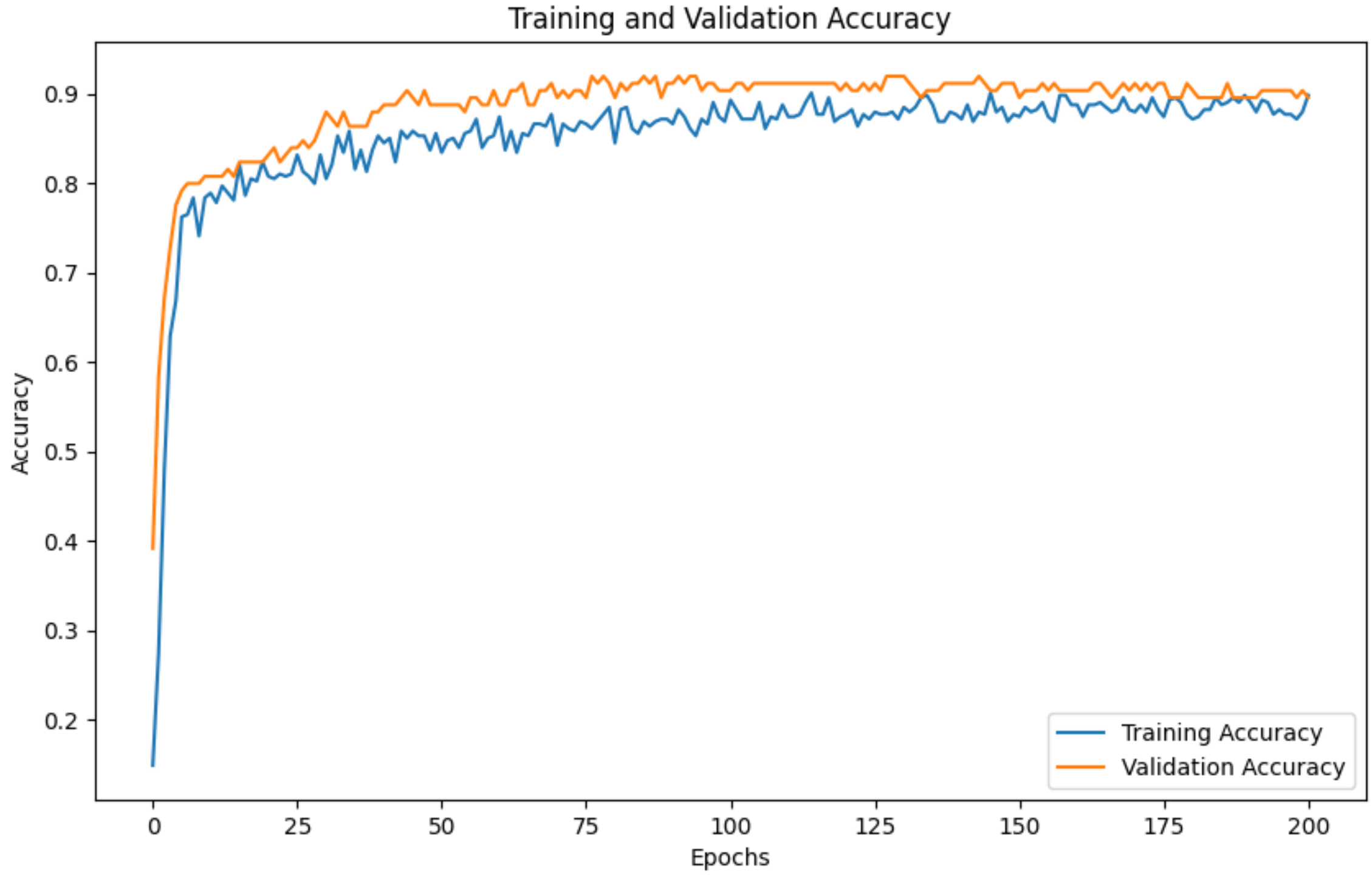

*Figure 7 Training and validation accuracy graph for the Set 3 training experiment*

The model's performance across the three gene sets shows a progression from moderate to high accuracy and a corresponding decrease in loss from high to low. This change through the different gene sets proves the importance of the gene set's quality and characteristics in training an effective model. While Set 1 and Set 2 provide low to moderate results, Set 3 displays excellent performance in comparison, demonstrating that it contains the most informative and useful features for the model to learn from.

## 3.2 Explainable AI Findings

This section describes the results of the three explainable AI experiments and the interpretation of the resulting SHAP values for each trained model. The SHAP value plots, which can be observed in Figures 9, 10, and 11, represent the average impact of different genes on the output of our predictive models for HCC stages.

Figure 8 represents the SHAP plots for the PSO graph, whereas Figure 9 represents the genetic algorithm graph and Figure 10 shows the ant colony graph. The x-axis in each figure shows the mean absolute SHAP value, which represents the average contribution of each gene to the model's predictions, whereas the y-axis lists the genes used as features from the dataset. Higher SHAP values mean that the gene has a stronger impact on the model's decision, and the coloured segments within each bar represent the impact of the gene on predicting different HCC development stages (or classes), such as the ehcc stage being associated with blue or phcc with purple.

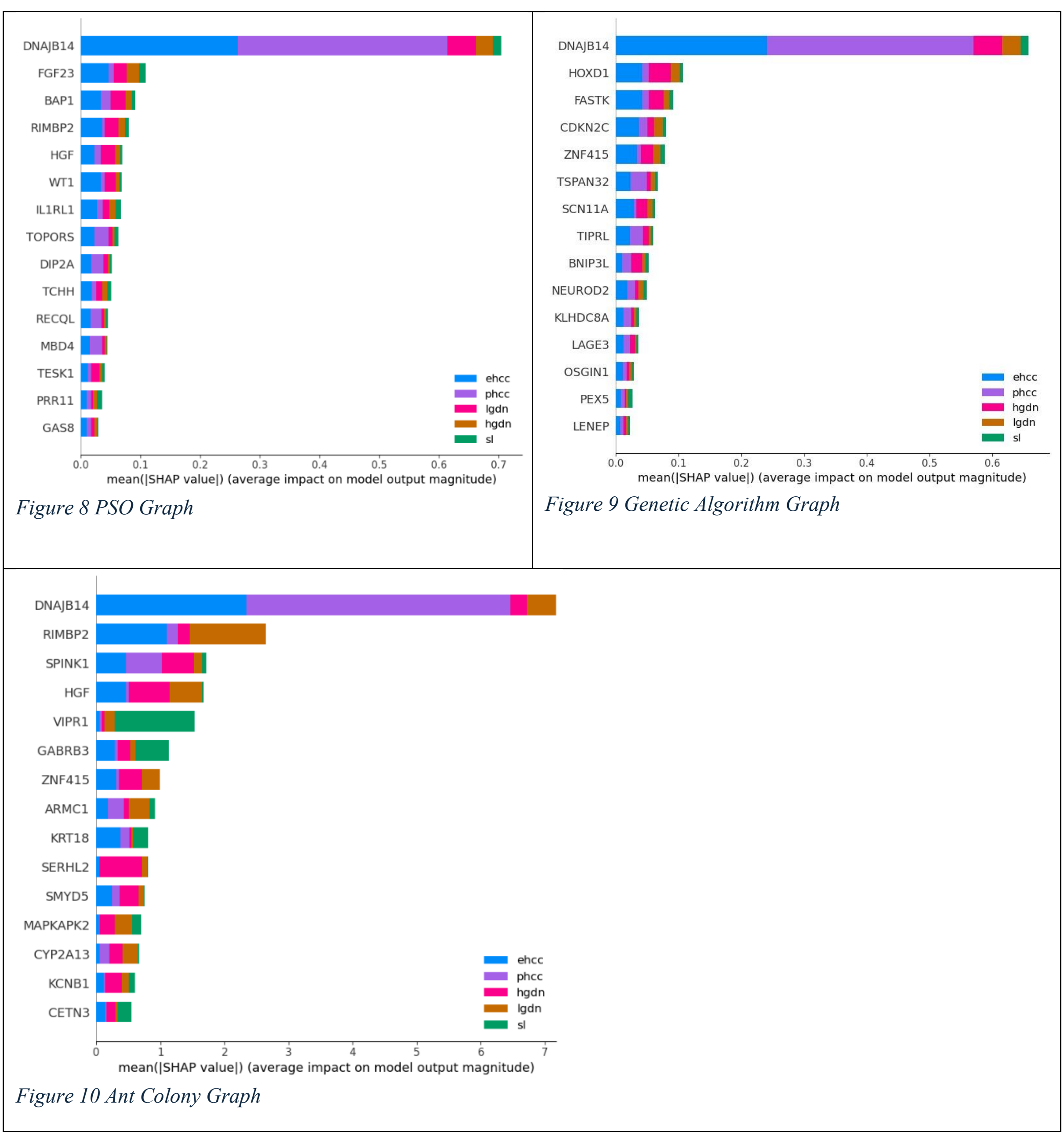

*Figure 8 PSO Graph*

*Figure 9 Genetic Algorithm Graph*

*Figure 10 Ant Colony Graph*

**DNAJB14** can be observed to be the most influential gene in all three models, with a consistently high SHAP value across different hepatocarcinogenesis stages. DNAJB14's contribution is primarily towards predicting the ehcc and phcc classes, suggesting its importance in distinguishing these stages. Other genes show a much more balanced influence across different stages, and their relative importance differs between models, because different optimization approaches (Genetic Algorithm, PSO, and Ant Colony Optimization) can end up selecting different optimal features with varying levels of contribution. **DNAJB14** appears as the top feature in all models, and **ZNF415** appears in the graphs belonging to Figure 9 and Figure 10, whereas **RIMBP2** and **HGF** appear in the graphs belonging to Figure 8 and Figure 10.

The third model in Figure 10 appears to produce significantly higher mean SHAP values compared to the two other models, as observed in its x-axis scale, suggesting that it might have identified genes with a stronger contribution to classification. The first two models (Genetic

Algorithm and PSO) have lower but more evenly distributed SHAP values across genes as observed in Figure 8 and Figure 9, indicating that they may rely on a more balanced set of features compared to the ant colony-based model. Despite the prevalence of **DNAJB14** across all graphs, it has not been previously reported to be involved in HCC in the existing literature, and hence further clinical validation and research on possible links between this gene and HCC are warranted.

## 3.3 Clinical Validation Findings

DNAJB14, also known as heat shock protein 40kDa (HSP40), belongs to a group of DNAJ proteins which are localized in the endoplasmic reticulum (ER) and are essential co-chaperones of heat shock protein 70 (Hsp70) [56]. The primary function of this protein is to accelerate the proteasome-dependent degradation of misfolded transmembrane proteins, particularly under condition of ER stress. [57] Studies have shown that DNAJB12 and DNAJB14 can interact in cancer cells to facilitate ER protein reflux into the cytosol and promote pro-tumourigenic features. [58] Further evidence suggests that it may serve as a drug target in non-endometrioid carcinoma, while in HCC it may affect patient survival by regulating the tumour microenvironment through recruitment of NK cells [59] [60]. Yet, further evidence on the role of DNAJB14 in liver cancer is very limited. To evaluate whether the identified gene DNAJB14 has functional implications in HCC, *in vitro* validation was performed as shown in Figure 11. First, the association of DNAJB14 with overall survival in HCC patients was shown to be significant in public datasets. Here, the patients with high expression exhibited worse survival than the patients with low expression (Figure 11a). To further evaluate its role on HCC cells, siRNA-mediated knockdown of DNAJB14 was effectively performed, as confirmed by qRT-PCR and Western blotting (Figure 11b+c). Increasing concentrations of siRNA against DNAJB14 significantly affected cell viability, resulting in increased cell death, with highly pronounced effects at 150 and 200 pM siRNA. This confirmed its relevance in HCC cell survival (Figure 11d). A further effect of DNAJB14 knockdown on cell proliferation and self-renewal was demonstrated using colony and sphere formation assays (Figure 11d+e). A significant reduction in colony and sphere formation capacity was observed, with up to 52% reduction in colonies and 48% reduction in spheres. Finally, the effects on migratory and invasive properties were tested, where DNAJB14 knockdown reduced cell migration, resulting in slower gap closure and lower invasion rate, as reflected by a lower cell index (Figure 11f+g). Overall, these results demonstrated the important role of DNAJB14 in HCC, validating the findings from artificial intelligence and machine learning approaches.

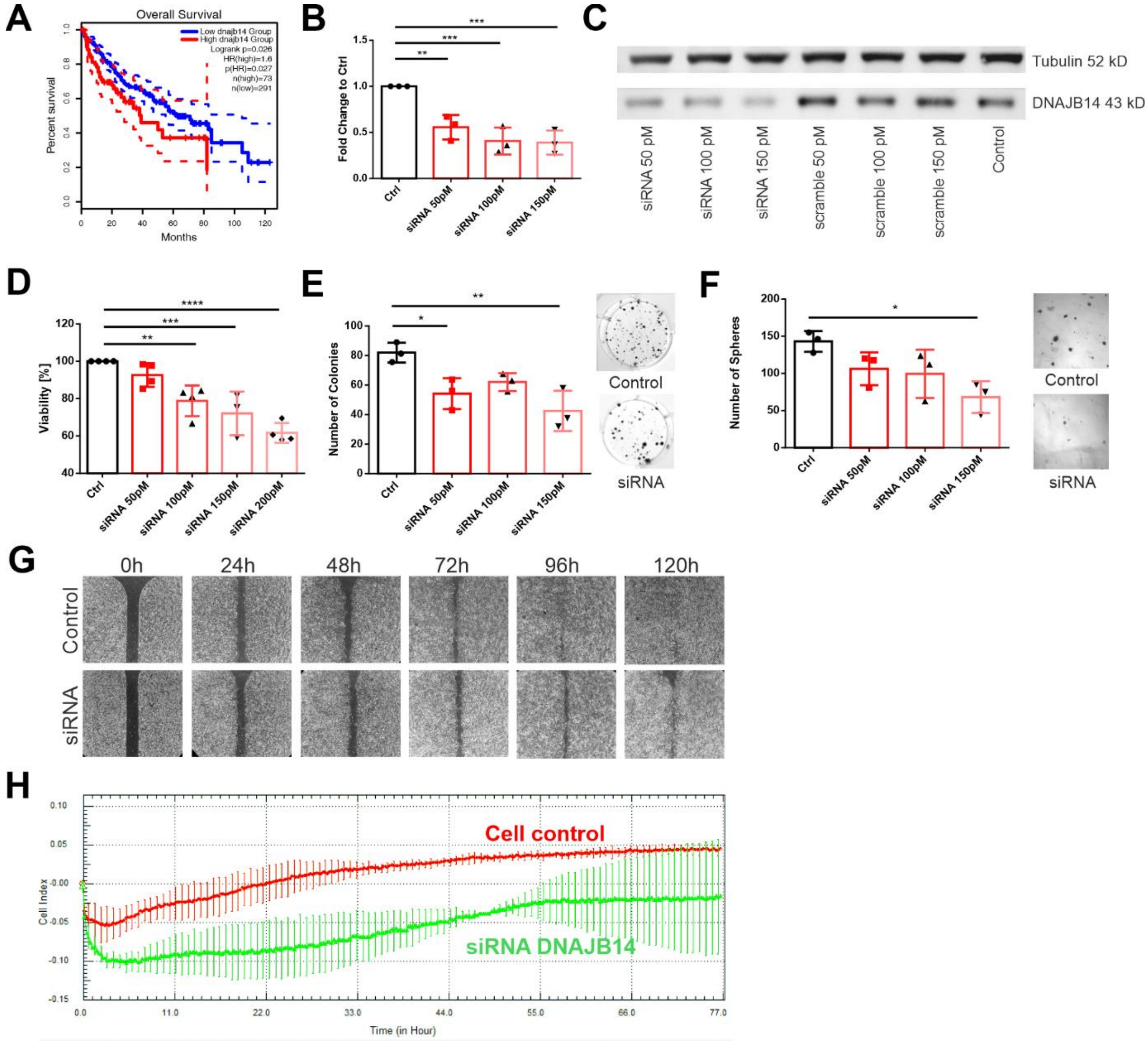

*Figure 111 In vitro functional validation of DNAJB14*

(A) Kaplan-Meier curve representing the overall survival of HCC patients with high and low expression of DNAJB14. The data was obtained from public dataset (TCGA) (B) Gene and (C) protein expression of DNAJB14 after treatment with increasing concentrations of DNAJB14 siRNA for 72 h. (D) Cell viability assay after treatment with increasing concentrations of DNAJB14 siRNA. (E+F) Representative images of colony and sphere formation assay and number of colonies and spheres after siRNA-mediated DNAJB14 knockdown. (G) Wound-healing assay showing the migratory properties of the cells. Gap closure reflects the speed and ability of cells to effectively migrate. (H) Invasion assay curves measured by xCELLigence represent the real-time observation of cell invasion through collagen I and IV layers on CIM-plate 16. The cell index reflects the number of invading cells. Mean±SD, n=3, *p<0.05, **p<0.01, ***p<0.001, ***p<0.0001

# 4. Conclusions

This research has investigated the application of deep learning and explainable artificial intelligence in liver cancer diagnosis for five different stages of hepatocarcinogenesis based on a genomic biomarkers HCC dataset constructed using semi-supervised learning from three source datasets. Several deep learning experiments were conducted using different feature extraction techniques and gene sets to determine the best features for training a model with the highest possible accuracy and minimal loss metrics, in addition to using class weights to overcome the class imbalance in the training data. The highest accuracy model was achieved using 15 selected genes with the SelectKBest algorithm, at 90.74% accuracy, and the lowest model loss was recorded using 20 genes at 0.3187. By using various optimization algorithms and classification models in addition to SHAP, this study has also demonstrated how XAI can provide insights into model decision-making in the context of liver cancer diagnosis and prediction with a high degree of accuracy in classifying liver cancer, identifying DNAJB14 as the most consistently influential gene across all three SHAP graphs.

The main limitation in this study is the class imbalance in the dataset, with the 'sl' class having significantly fewer samples compared to the 'phcc' and 'ehcc' classes. While a weighted training approach was used to mitigate this, more research and data are needed to mitigate the negative effects of this imbalance on the model's performance and generalizability. For future work, genetic variations, environmental factors, and other clinical differences among patients from different regions can also affect the model's accuracy when applied to new populations, which means that genomic expression data may not be the only contributing factor in liver cancer development and progression, requiring future research to explore combinations with other features and patient data. To investigate whether the identified gene across all three SHAP graphs is clinically relevant, functional in vitro validation was performed in the HCC model. Analysis of DNAJB14 inhibition clearly showed a significant impact on cellular properties. This was strongly reflected in reduced cell survival, self-renewal and impaired pro-metastatic properties with reduced cell migration and invasion. Even though this gene has not been previously reported to be highly associated with HCC in the literature, it was identified as the most influential gene in all three explanatory AI models and was successfully validated in HCC for the first time.

**Ethics Statement:** The authors would like to convey their thanks and appreciation to the University of Sharjah and University of Lubeck for supporting the work. All experiments were conducted on publicly available datasets
**Consent for publication:** Not Applicable
**Competing Interests:** "The authors declare that they have no conflict of interest."
**Informed consent:** "This study does not involve any experiments on animals."
**Data Availability:** The datasets used in this research are publicly available and are described in Section 2.1.
**Funding:** This project was funded by University of Sharjah and University of Lubeck